\documentclass{article}

\PassOptionsToPackage{numbers, compress}{natbib}
\usepackage[preprint]{neurips_2023}
\usepackage{floatrow}
\newfloatcommand{capbtabbox}{table}[][\FBwidth]

\usepackage[utf8]{inputenc} % allow utf-8 input
\usepackage[T1]{fontenc}    % use 8-bit T1 fonts
\usepackage[hidelinks]{hyperref}       % hyperlinks
\usepackage{url}            % simple URL typesetting
\usepackage{booktabs}       % professional-quality tables
\usepackage{amsfonts}       % blackboard math symbols
\usepackage{nicefrac}       % compact symbols for 1/2, etc.
\usepackage{microtype}      % microtypography
\usepackage{xcolor}         % colors
\usepackage{graphicx}
\usepackage{amsmath}
\usepackage{amsthm}
\usepackage{tcolorbox}
\usepackage{caption}
\usepackage{listings}% http://ctan.org/pkg/listings
\title{A New Approach Towards Autoformalization}

\newcommand{\dataset}[0]{\normalfont\textbf{arXiv2Formal}}

\author{%
  Nilay Patel \\
  Department of Computer Science\\
  University of California, Santa Cruz\\
  \texttt{nilay@ucsc.edu} \\
  \And
  Rahul Saha \\
  \texttt{rsaha@alumni.princeton.edu} \\
  \AND
  Jeffrey Flanigan \\
  Department of Computer Science \\
  University of California, Santa Cruz \\
  \texttt{jmflanig@ucsc.edu} \\
}

\begin{document}

\maketitle

\begin{abstract}
Verifying mathematical proofs is difficult \cite{poincare-conjecture-article}, but can be automated with the assistance of a computer. Autoformalization is the task of automatically translating natural language mathematics into a formal language that can be verified by a program. This is a challenging task, and especially for higher-level mathematics found in research papers. Research paper mathematics requires large amounts of background and context. In this paper, we propose an avenue towards tackling autoformalization for research-level mathematics, by breaking the task into easier and more approachable subtasks: unlinked formalization (formalization with unlinked definitions and theorems), entity linking (linking to the proper theorems and definitions), and finally adjusting types so it passes the type checker. In addition, we present \dataset,\footnote{\url{https://github.com/jlab-nlp/arxiv2formal}} a benchmark dataset for unlinked formalization consisting of 50 theorems formalized for the Lean theorem prover \cite{lean-theorem-prover} sampled from papers on \url{arXiv.org}.
\end{abstract}

\section{Introduction}
Mathematics is a notoriously difficult, and introducing even one false theorem can invalidate all follow-up work. Discounting the time it takes to prove results, it often takes many months or years to verify the result and have it accepted by the community. Still, mistakes can be made and even seemingly simple ones can be difficult to find. One famous example is a mistake in knot theory which went uncorrected for nearly 100 years.\footnote{\url{https://mathoverflow.net/a/9059}}

Using proof assistants to verify proofs is effective, but requires the mathematics to be written in a \textit{formal language} as opposed to the \textit{natural language} humans use. Formalizing natural language mathematics has a high barrier to entry, even for mathematicians, and remains difficult and time-consuming once learned. To mitigate these high costs, research in automatic formalization (\textit{autoformalization}) and automatic theorem proving has been increasingly popular in recent years \cite{szegedy-path,zheng2021minif2f, autoform-with-llm, towards-formalization-with-llms, proofnet, draft-sketch-prove, leandojo, formal-abstracts}, in part due to major advances in natural language processing (NLP) such as the introduction of large language models (LLMs) \cite{lms-are-fewshot-learners}. A high-quality autoformalization tool would alleviate many downsides of theorem provers while maintaining their benefits.

However, there are still many challenges faced in the development of such an autoformalization tool, such as the intrinsic difficulty of the domain and severe lack of high-quality data. Previous work \citep{proofnet, leandojo,zheng2021minif2f, math-dataset, archive-of-formal-proofs} introduce datasets for this and other related tasks, but these works either are not parallel corpora needed for building a supervised model, or focus on "easier" domains, such as olympiad-style or undergraduate mathematics.\footnote{These domains are still unsolved and are by no means easy, but pose fewer challenges than higher-level mathematics.} While a useful stepping stone, they are far removed from "real world" mathematics which requires far more context and background material for theorems. One attempt at this task was Formal Abstracts \cite{formal-abstracts}, a project intending to formalize the main theorem of papers from arXiv, but progress on this project has ceased, which we speculate is partly due to the monumental challenge straight formalization of high-level mathematics poses (see \autoref{sec:related-work}).

Another looming issue with previous approaches to autoformalization is scalability. Much of the difficulty in formalization comes from the necessity for the foundations of the theory to be general and broadly applicable. Formalization for well-understood areas of mathematics can take years to develop; some areas of undergraduate mathematics remain incomplete \cite{mathlib}. Even a single theorem in high-level mathematics can take years to properly formalize \cite{castelvecchi2021mathematicians}.\footnote{It took a team over 1.5 years to formalize the prove a single theorem in the Liquid Tensor Experiment \cite{castelvecchi2021mathematicians, liquid-tensor-challenge, liquid-tensor-completion}.} Current work on autoformalization assumes that definitions and types are decided ahead of time by annotators, which is burdensome and potentially inhibitory to large-scale annotation projects.

%Much of the difficulty in formalization is deciding the types of various mathematical objects, because the objects will be re-used in many places. Understanding how the types of all the objects will interact in even a small area of mathematics can be very complex and take many years to work out.\jmf{add citation} %, even for seemly simple areas of mathematics.
%Current work on autoformalization assumes that the types are decided ahead of time by the annotators, which puts a large burden on the annotators, and is potentially not scalable to large annotation efforts.

%Often in mathematics, we break difficult problems down into easier tasks and build up a solution in pieces.
To tackle higher-level mathematics, we propose breaking the task of autoformalization down into easier subtasks to better leverage the powerful tools from NLP. Furthermore, we release a beta version of \dataset, a benchmark dataset for the first subtask, with 50 examples (and growing). For this work, we use Lean 3 \cite{lean-theorem-prover}, but a similar technique could be adapted for others. We also plan to release a version of the dataset in Lean 4.

At a high level, our subtasks to autoformalization are as follows:
\begin{enumerate}
    \item \textbf{Formalize Theorems (Unlinked)}: Translate natural language theorem statements into syntactically correct but not necessarily semantically valid Lean. In this intermediate step, we allow references to variables, functions, or other theorems which don't yet exist.
    \item \textbf{Formalize Definitions (Unlinked)}: Translate definitions of functions, variables, etc. from a paper into a "library" which can be referenced by theorems. These definitions should be recursively constructed using a base library (e.g., Lean's mathlib \cite{mathlib}).
    \item \textbf{Entity Linking}: Link referenced names to the definitions in the library from step 2. Since steps 1 and 2 generate names automatically, this step is necessary to resolve coreferences.
    \item \textbf{Adjust Types}: Postprocess the generated translations with the assistance of the Lean type checker to ensure types are aligned. This also may involve adjusting type signatures to add arguments to definitions which, in text, don't appear to need any, but are necessary in a formal setting.
\end{enumerate}

Note that theorem proving poses its own (separate) challenges compared to theorem formalization. For this work, we focus on the previous four tasks and leave theorem proving for future work.

The rest of the paper is organized as follows. First, we discuss some related work (\autoref{sec:related-work}), then we describe our proposed approach and reasoning (\autoref{sec:approach}), and finally we provide a benchmark dataset for the first subtask (\autoref{sec:dataset}).

\begin{figure}
    \centering
    \includegraphics[width=\textwidth]{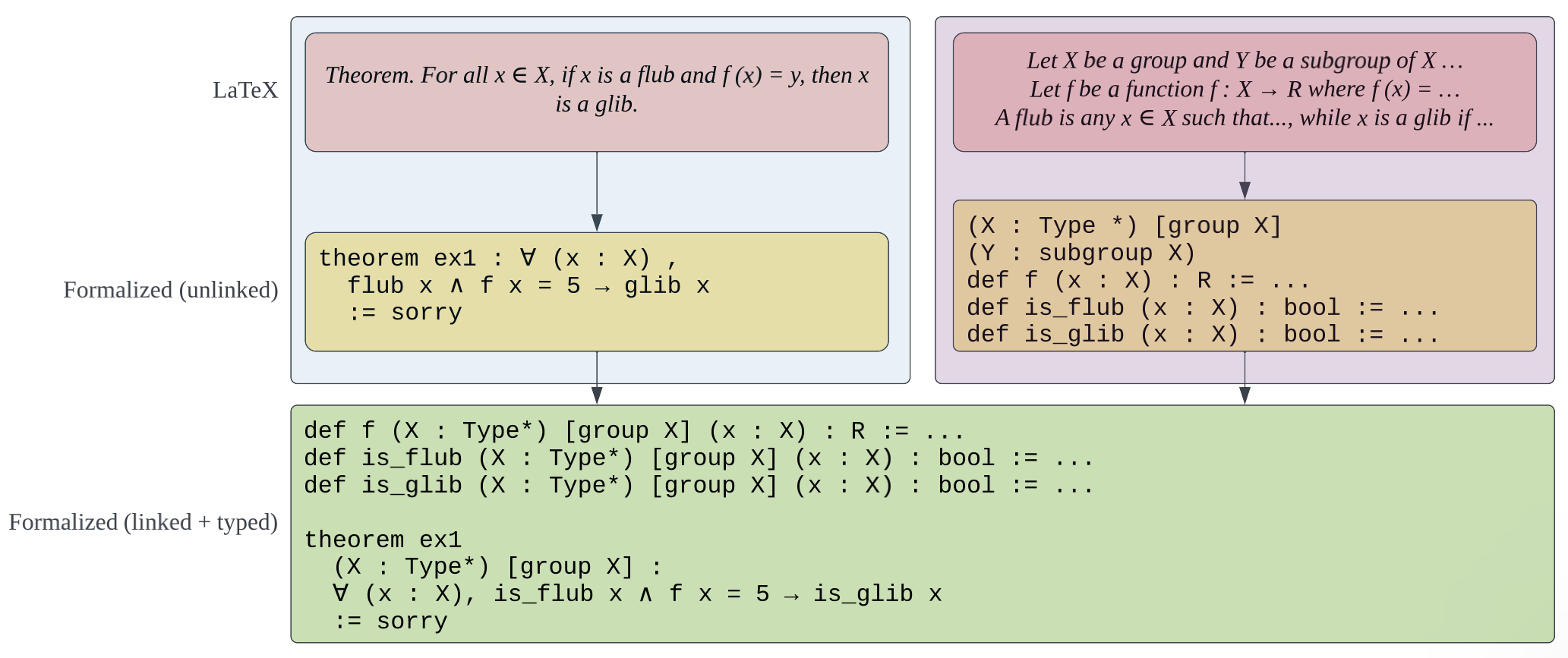}
    \caption{An example of our proposed autoformalization process, starting from LaTeX theorems (top left red) and definition (top right red) and ending with formalized Lean code (bottom). In step 1 (top left), despite lacking context for the set $X$, the definitions of flubs and glibs, and the function $f$, we are still able to make a plausible translation which we can fill in later. In step 2 (top right), we formalize definitions and determine the correct types of objects. Note that the translations are still not correct, since the type of X hasn't been applied to the other definitions or theorems yet. In step 3 (bottom), we fix this by linking the referenced name \textit{X} with the generated type \texttt{X}, and we also align the names of other mismatched functions.} 
    \label{fig:method}
\end{figure}

\section{Related Work}
\label{sec:related-work}
Several papers have been published recently in autoformalization and theorem proving \cite{archive-of-formal-proofs, math-dataset, proofnet, zheng2021minif2f, draft-sketch-prove, proof-artifact-cotraining}, but current research on autoformalization is limited by a small amount of high-quality data. Of note, miniF2F \cite{zheng2021minif2f} contains 488 formalized olympiad problems in Lean and three other languages. Codex \cite{codex} and other LLMs have been effective in autoformalization for this dataset, achieving over 25\% exact translation accuracy \cite{autoform-with-llm}. 

ProofNet \cite{proofnet} is another dataset consisting of 371 formal/informal undergraduate math problems in Lean 3. In addition to the dataset, the authors also describe two techniques which improve LLM autoformalization and theorem proving performance, prompt retrieval and distilled backtranslation. Other work with applying LLMs to autoformalization \cite{autoform-with-llm,towards-formalization-with-llms,towards-assistant-with-llm} have shown promising results. The authors of ProofNet also released an LLM pretrained on the proof-pile \cite{proofnet}, a large corpus of math theorems and proofs, which may further improve performance of autoformalization with LLMs.

Interestingly, an error analysis from \cite{autoform-with-llm} showed that a majority of the autoformalization errors on the minif2f dataset were due to an inability to link natural language and formal definitions. In our proposed task, this failure case would be handled by the linking in step 3, which should significantly improve overall performance.

\section{Our Approach to Autoformalization}
\label{sec:approach}

We propose to break autoformalization into four subtasks, each of which are easier than the original task.
\begin{enumerate}
    \item \textbf{Formalize Theorems (Unlinked).} Previous work \citep{autoform-with-llm} has shown promising results in the use of LLMs for autoformalization. A major problem, however, is high-level mathematics often requires a large amount of context even for short theorems.  %Oftentimes, there are many definitions, of which only a subset are used in any one theorem. 
    To simply the problem, we propose separating the tasks of (1) translating theorems to a formal language with placeholder names and (2) linking the placeholders to the actual definitions. See \autoref{fig:method} for an example.
    \item \textbf{Formalize Definitions (Unlinked).} Similar to theorems, definitions often require prior context and thus need to be constructed iteratively. However, even without context, we can use the same approach as above and formalize the definition with placeholder names.
    \item \textbf{Link Entities.} Unlinked formalization in steps 1 and 2 result in list of theorems and definitions, but the identifiers are not necessarily aligned. %For example, a theorem may reference a vector space without specifying the field, but mathlib defines a vector space as a module over a field and this mismatch can cause problems.
    For example, a theorem may reference a vector space, but mathlib does not define vector spaces, and instead uses a module over a field.  The reference to a vector space needs to be linked to mathlib's module definition.
    While still challenging, entity linking and coreference resolution is a well-studied field in NLP, and we plan to adapt and leverage prior work for this task.
    \item \textbf{Adjust Types.} This last step is a catch-all for any processing that requires the assistance of the Lean type system, such as correcting type misalignment. In practice, what this step entails is dependent on the output of the previous three steps. As an example, if a theorem references a vector space without specifying the field, but we linked it to mathlib's module definition, which requires a specified field, we would need to add the field argument to correct the types.
    %As the previous three steps get closer to being solved, postprocesssing becomes less necessary, though may still useful for verification.
\end{enumerate}
%\paragraph{Formalize Theorems} 
%\paragraph{Formalize Definitions}
%\paragraph{Entity Linking}
%\paragraph{Postprocess}

% \begin{figure}[t]
%   \begin{tcolorbox}[minipage,arc=0pt, standard jigsaw, opacityback=0, outer arc=0pt, size=normal]
%     \begin{tabular}[t]{l l}
%          \textbf{Natural Language} & \textbf{Theorem.} \textit{For all $x \in X$, if $x$ is a flub and $f(x) = y_0$, $x$ is a glib.} \\
%          \midrule
%          \textbf{Formalized (Lean 3)} &
% \begin{lstlisting}
% theorem ex1 (X : Type*) [group X] : 
%   $\forall$ (x : X), 
%   is_flub x $\wedge$ f x = 5 $\to$ is_glib x := sorry
% \end{lstlisting}
%     \end{tabular}
%   \end{tcolorbox} 
%   \caption{Now that we've linked the definition of $X$ with its reference in the theorem, we are able to incorporate it into the theorem hypothesis. Similarly, we've aligned \texttt{is\_flub} and \texttt{is\_glib} to their counterparts \texttt{flub} and \texttt{glib}.}
%   \label{fig:example-step-3-and-4}
% \end{figure}

\section{Benchmark Dataset: \dataset}
\label{sec:dataset}

To facilitate progress towards subtask 1, we release a benchmark dataset of 50 theorems in LaTeX and Lean 3 sampled from a random subset of mathematics papers from \url{arXiv.org}. On average, our natural language theorems consisted of 140 tokens compared to 113 tokens for the formalized versions. Currently, \dataset~is a beta release and will be updated regularly as more examples are annotated. For examples, see \autoref{app:examples}.

We construct the dataset in a semi-automated manner.  We formalize the theorems first automatically via prompting GPT-3.5 turbo \cite{chatgpt} and manually correcting any mistakes in the automatic translation. As more translations are completed, we use them to improve the quality of future translations by selecting them as examples for the prompt. We observe that the amount of human correction work lessens as the dataset becomes larger because the prompts become better (\autoref{tab:gpt-performance}).

%\subsection{Statistics}
%\label{subsec:statistics}
% \paragraph{Statistics}\label{par:statistics} For the beta release, we have 50 examples from 16 papers across a variety of topics, which we bucket into five broad categories seen in \autoref{fig:topic-distribution}. On average, natural language theorems were approximately 140 tokens in length, whereas the formalized version were only 113 tokens on average---nearly a 20\% decrease in length.

\begin{minipage}{0.45\textwidth}
  %\begin{figure}[h]
      \centering
      \includegraphics[width=\textwidth]{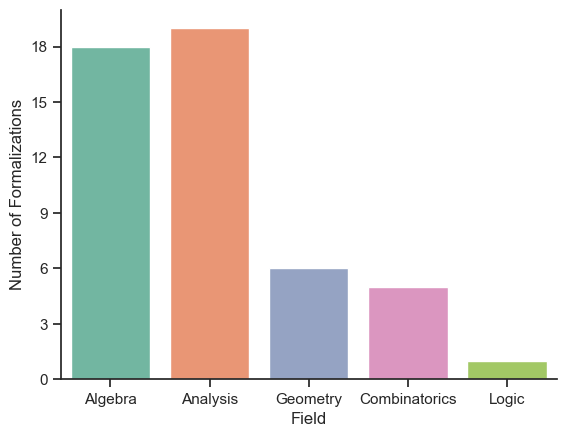}
      \captionof{figure}{Distribution of topics of formalized theorems in \dataset.}    \label{fig:topic-distribution}
  %\end{figure}
 \end{minipage}
 \hfill
 \begin{minipage}{0.5\textwidth}
%  \begin{table}[h]
      \centering
    \begin{tabular}{llll}
        \toprule
        Model & BLEU $\uparrow$ & TER $\downarrow$ & TrueSkill $\uparrow$ \\
        \midrule
        GPT-3.5 & 17.2 & 140.5 & 10.9 \\
        +ICL 1 & 31.9 & 93.4 & 25.4 \\
        +ICL 5 & 37.5 & 85.7 & 28.0, \\
        +ICL 10 & \textbf{41.7} & \textbf{83.9} & \textbf{33.1} \\
        \bottomrule
    \end{tabular}
        \captionof{table}{Performance of GPT-3.5 turbo (16k) on \dataset~with varying number of examples for in-context learning (ICL). Adding just a few examples to show the model the expected format significantly improves BLEU score, showing promising results for this subtask.}
    \label{tab:gpt-performance}
 % \end{table}
 \end{minipage}

\paragraph{Evaluation} We evaluate using common NLP automatic metrics such as BLEU \cite{bleu} and Translation Error Rate (TER) \cite{ter}, as well as human evaluations. In a small-scale experiment of 10 examples, we compared model outputs with varying BLEU and Translation Error Rate (TER) scores, and found that an increase in these scores strongly correlates with better human evaluation of the output ($R=.99$). For the future, we suggest evaluation with automatic metrics combined with a human evaluation.

\subsection{Experiments with GPT-3.5}
\label{subsec:human-analysis}

We experimented with unlinked-autoformalization for \dataset~using GPT-3.5 \cite{chatgpt}. GPT-3.5 likely contains training data from mathlib, and without any prompting, it formalizes statements in a very similar way to theorems found there. With few-shot in-context learning, performance on \dataset~increased noticeably (\autoref{tab:gpt-performance}).  %This is extremely beneficial in improving baseline translation quality, as most generated Lean is syntactically valid and in a familiar structure. In particular, GPT-3.5 was effective at translating equations from LaTeX to Lean3, though performance would diminish with unusual notation and long theorems.

GPT-3.5 was highly effective in assisting our formalization. However, the naive method we used for these baseline results made several mistakes. One common mistake is that GPT often attempts to guess the type of referenced objects based on insufficient context. While it was sometimes correct (e.g., seeing familiar vocabulary and LaTeX formatting, it guesses that $V$ is a vector space), it hallucinates regularly and mistypes in most cases.  In our proposed autoformalization approach, these errors will be fixed later when more context is available to adjust types in step 4.

One of the more difficult cases which GPT struggled to handle is multipart theorems. In many cases, each part requires different or mutually exclusive conditions which makes formalizing it as a single theorem awkward and unwieldy. A much more natural approach is to split the theorem into multiple parts, though this introduces another challenge of automatic theorem splitting. %as relevant data is even more limited.

\section{Conclusion}
\label{sec:conclusion}
Previous work in autoformalization largely focuses on "easier" domains such as olympiad or undergraduate math. Higher-level mathematics requires far more background and context for each theorem and thus requires a new approach. We propose breaking the task down autoformalization into four more manageable subtasks: (1) formalizing theorems (unlinked), (2) formalizing definitions (unlinked), (3) linking identifiers and references, and (4) adjusting types. While still difficult, each subtask is much easier to tackle in isolation than the full task and allows for cheaper data collection. We hope to stimulate research along this path, and provide \dataset, a dataset of 50 machine-assisted human-annotated LaTeX/Lean 3 theorems as a benchmark for subtask 1. We welcome any contributions to future versions of this dataset.

%\bibliographystyle{authoryear}
%\bibliography{leanify}

% (nilay) I used biblatex here instead of natbib since I couldn't get it to work right, we should probably fix this before submitting.
\bibliography{leanify}
\bibliographystyle{plainnat}
\clearpage
\appendix

\section{Appendix: Dataset Examples}
\label{app:examples}

\begin{figure}[hb]
  \small
  \begin{tcolorbox}[minipage,arc=0pt, standard jigsaw, opacityback=0, outer arc=0pt, size=normal]
    \begin{tabular}{l l}
     \textbf{Natural Language} & \textbf{Lemma.} \textit{For any irrational number $x$, $x$ is a Brjuno number } \\ 
     & \textit{if and only if $x$ and $-x$ are semi-Brjuno numbers.} \\
     \midrule
    \textbf{Formalized (Human)} &
\begin{lstlisting}
lemma theorem_two
  (x : $\mathbb{R}$) (hx : $\lnot$ is_rational x) :
  is_brjuno x $\leftrightarrow$ 
  (is_semi_brjuno x $\wedge$ is_semi_brjuno (-x))
  := sorry
\end{lstlisting} \\
\midrule
\textbf{Formalized (GPT-3.5)} &
\begin{lstlisting}
lemma theorem_two
  (x : irrational_number) :
  is_brjuno_number x $\leftrightarrow$ 
  (is_semi_brjuno_number x $\wedge$ 
   is_semi_brjuno_number (-x))
  := sorry
\end{lstlisting} \\
    \end{tabular}
  \end{tcolorbox} 
  \caption{An example pair from an arXiv paper (\cite{arxiv-example-0705.1690}) from the benchmark dataset and its formalized versions. Note that we don't know the definitions of semi-Brjuno and Brjuno, but we are still able to formalize the theorem with placeholders.}
  \label{fig:example-from-dataset-1}
\end{figure}

\begin{figure}[htb]
  \small
  \begin{tcolorbox}[minipage,arc=0pt, standard jigsaw, opacityback=0, outer arc=0pt, size=normal]
    \begin{tabular}{l p{8.5cm}}
     \textbf{Natural Language} & \textbf{Lemma.} \textit{Let $G$ be a finitely generated group with finite generating set $S$ and the associated word length metric. Then the following are equivalent:
\begin{enumerate}
\item $G$ is amenable;
\item For any $\varepsilon>0$, there exists a function $f\in \ell^2(G)_{1,+}$ such that $\|f-s\cdot f\|_2\leq \varepsilon$ for any $s\in S$ and $\mathbf{supp}(f)$ is finite.
\end{enumerate}}\\
     \midrule
    \textbf{Formalized (Human)} &\begin{lstlisting}
lemma prop_hulanicki_reiter
  (G : Type*) [group G]
  (S : finset G)
  (hS : finite_generating_set S)
  (word_length : word_length_on G S) :
  amenable G $\leftrightarrow$
    ($\forall$ ($\varepsilon$ : $\mathbb{R}$) (h$\varepsilon$ : $\varepsilon$ > 0),
      $\exists$ (f : G $\to \mathbb{R}$) (hf : f $\in l^2G$) 
        (hf_pos : $\forall$ (g : G), g $\neq$ 1 $\to$ f g > 0),
        ($\forall$ (s : G), s $\in$ S $\to$ 
          $\|$f - ($\lambda$ (g : G), f (s $*$ g))$\|_2$ $\leq \varepsilon$)
          $\wedge$ (set.finite (function.support f)))
  := sorry
\end{lstlisting} \\
\midrule
\textbf{Formalized (GPT-3.5)} &
\begin{lstlisting}
lemma prop_hulanicki_reiter
  (G : Type*) [group G]
  (S : finset G)
  (hS : finite_generating_set S)
  (word_length : word_length_on G S) :
  (amenable G) $\leftrightarrow$
    ($\forall$ ($\varepsilon$ : $\mathbb{R}$) (h$\varepsilon$ : $\varepsilon$ > 0),
      $\exists$ (f : G $\to \mathbb{C}$) (hf : f $\in$ l2_1_plus G) 
        (hf_pos : $\forall$ (g : G), g $\neq$ 1 $\to$ f g > 0),
        ($\forall$ (s : S), 
          norm_l2 (f-(s*f)) $\leq \varepsilon$)
          $\wedge$ (finite_support f))
  := sorry
\end{lstlisting} \\
    \end{tabular}
  \end{tcolorbox} 
  \caption{An example pair from an arXiv paper (\cite{arxiv-example-0705.1690}) from the benchmark dataset and its formalized versions. In this more complicated case, GPT is still able to correctly translate several pieces of the theorem. One mistake is the fact that $f$ should have a range of $\mathbb{C}$ rather than $\mathbb{R}$, but this is a mistake that will be corrected in stage 4 (entity linking). The use of $\|\ldots\|_2$ instead of our preferred \texttt{norm\_l2} is due to the prevalence of this notation in the pretraining corpus.}
  \label{fig:example-from-dataset-2}
\end{figure}

\end{document}